\def\year{2022}\relax
\documentclass[letterpaper]{article} 
\usepackage{aaai22}  
\usepackage{times}  
\usepackage{helvet}  
\usepackage{courier}  
\usepackage[hyphens]{url}  
\usepackage{graphicx} 
\usepackage{natbib}  
\usepackage{caption} 
\DeclareCaptionStyle{ruled}{labelfont=normalfont,labelsep=colon,strut=off} 
\usepackage{algorithm}
\usepackage{algorithmic}

\usepackage{comment}
\usepackage{subcaption}
\usepackage{microtype}
\usepackage{enumitem}
\usepackage{amsmath}
\usepackage{xcolor}
\usepackage{multirow}
\usepackage{amsfonts}
\usepackage{booktabs}
\usepackage{bbm}
\usepackage{pifont}
\usepackage{makecell,rotating}
    \setcellgapes{5pt}

\newcommand{\sy}[1]{\textcolor{magenta}{\bf\small [#1 -- SY]}}

\newenvironment{tightcenter}{%
  \setlength\topsep{0pt}
  \setlength\parskip{0pt}
  \begin{center}
}{%
  \end{center}
}

\makeatletter
\def\thickhline{%
  \noalign{\ifnum0=`}\fi\hrule \@height \thickarrayrulewidth \futurelet
   \reserved@a\@xthickhline}
\def\@xthickhline{\ifx\reserved@a\thickhline
               \vskip\doublerulesep
               \vskip-\thickarrayrulewidth
             \fi
      \ifnum0=`{\fi}}
\makeatother

\makeatletter
\newcommand\footnoteref[1]{\protected@xdef\@thefnmark{\ref{#1}}\@footnotemark}
\makeatother

\newlength{\thickarrayrulewidth}
\usepackage{newfloat}
\usepackage{listings}
\floatstyle{ruled}
\newfloat{listing}{tb}{lst}{}
\floatname{listing}{Listing}
\title{Retrieve, Caption, Generate: Visual Grounding for Enhancing Commonsense in Text Generation Models}
\author {
    Steven Y. Feng \textsuperscript{\rm 1}
    Kevin Lu        \textsuperscript{\rm 2}
    Zhuofu Tao        \textsuperscript{\rm 3}
    Malihe Alikhani \textsuperscript{\rm 4}
    Teruko Mitamura \textsuperscript{\rm 1}
    Eduard Hovy \textsuperscript{\rm 1}
    Varun Gangal  \textsuperscript{\rm 1}
}
\affiliations {
    \textsuperscript{\rm 1} Language Technologies Institute, Carnegie Mellon University 
    \textsuperscript{\rm 2} University of Waterloo \\
   \textsuperscript{\rm 3} University of California, Los Angeles
    \textsuperscript{\rm 4} School of Computing and Information, University of Pittsburgh\\
    {\tt syfeng@cs.cmu.edu,kevin.lu1@uwaterloo.ca,z24tao@g.ucla.edu,malihe@pitt.edu} \\
    {\tt teruko@cs.cmu.edu,hovy@cs.cmu.edu,vgangal@cs.cmu.edu} \\    
}

\begin{document}

\maketitle

\begin{abstract}
We investigate the use of multimodal information contained in images as an effective method for enhancing the commonsense of Transformer models for text generation. We perform experiments using BART and T5 on concept-to-text generation, specifically the task of generative commonsense reasoning, or \emph{CommonGen}. We call our approach \textit{VisCTG: Visually Grounded Concept-to-Text Generation}. VisCTG involves captioning images representing appropriate everyday scenarios, and using these captions to enrich and steer the generation process. Comprehensive evaluation and analysis demonstrate that VisCTG noticeably improves model performance while successfully addressing several issues of the baseline generations, including poor commonsense, fluency, and specificity.




\end{abstract}

\section{Introduction}
\label{sec:intro}
Transformer-based models have seen increasing popularity for NLP tasks and applications. This includes SOTA text generation models such as BART \cite{lewis-etal-2020-bart} and T5 \cite{JMLR:v21:20-074}. Larger corpora and better pretraining losses are major reasons driving these gains. However, despite increasing attention on the commonsense of models through works like COMET \cite{bosselut-etal-2019-comet}, studies have shown that even large pretrained models still struggle with commonsense tasks that humans can reason through very easily \cite{talmor2019olmpics}. We believe that there is commonsense information in other modalities like vision, beyond what is reported \cite{gordon2013reporting} in text, which can possibly augment commonsense and enhance decision-making processes of text generation models.

In this paper, we show this is true by improving the performance of Transformer-based text generation models on concept-to-text generation using visual grounding, which we call \textit{VisCTG: Visually Grounded Concept-to-Text Generation}. Concept-to-text generation is a high-level formulation of several constrained text generation and data-to-text natural language generation (NLG) tasks. These are challenging tasks that have seen increasing interest, and involve generating natural language outputs given certain pre-conditions, e.g. specific words in the outputs, and structured or semi-structured inputs. They typically involve converting a set of inputs into natural language text. These inputs can normally be thought of as \textit{concepts}, or high-level words or structures, that play an important role in the generated text. Multimodal work has seen increasing popularity, but its exploration for constrained and data-to-text NLG has been limited \cite{multimodal_survey_1,multimodal_survey_2}.\footnote{Code: \url{https://github.com/styfeng/VisCTG}}

\begin{table}[t]
\centering
\resizebox{\columnwidth}{!}{\begin{tabular}{ |c|c| } 
 \hline
 \textit{\{stand, hold, umbrella, street\}} & \textit{\{food, eat, hand, bird\}}\\
 \includegraphics[width=0.35\textwidth]{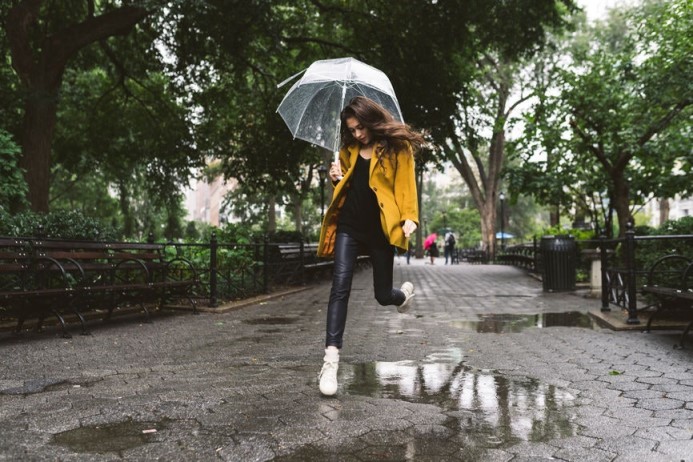} & \raisebox{1pt}{\includegraphics[width=0.35\textwidth]{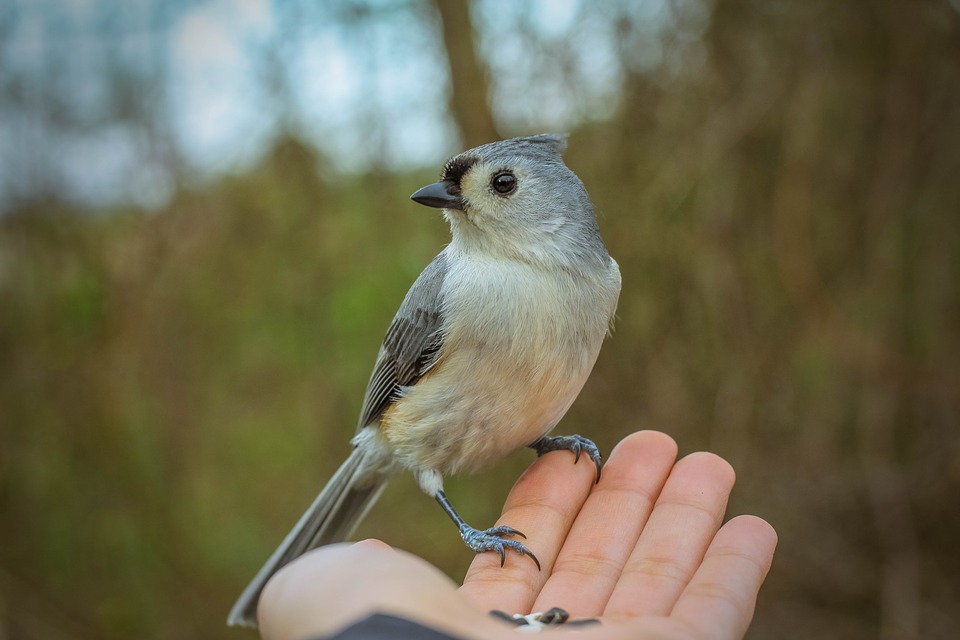}}\\
 \textbf{baseline:} A holds an umbrella while standing on the street & \textbf{baseline:} hand of a bird eating food\\
 \textbf{capt:} a woman walking down a street holding an umbrella & \textbf{capt:} a person holding a small bird in their hand\\
 \textbf{VisCTG:} A woman stands on a street holding an umbrella. & \textbf{VisCTG:} A bird eats food from a hand.\\[0.5mm]
 \hline
 \rule{0pt}{0.9\normalbaselineskip}
 \textit{\{cat, bed, pet, lay\}} & \textit{\{fence, jump, horse, rider\}}\\
 \includegraphics[width=0.25\textwidth]{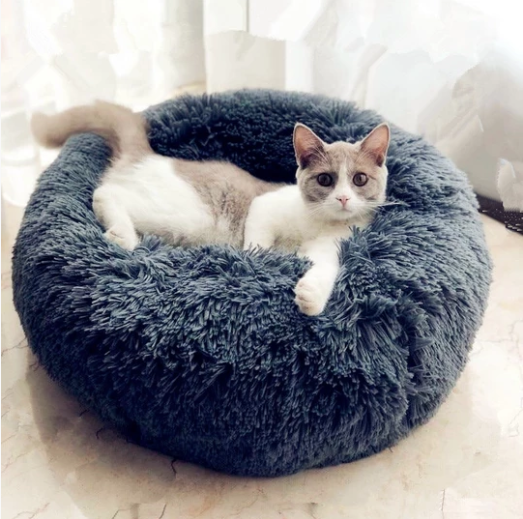} & \raisebox{1pt}{\includegraphics[width=0.35\textwidth]{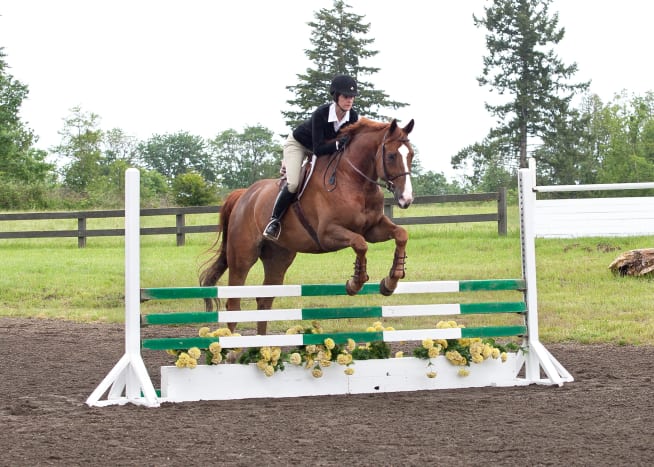}}\\
 \textbf{baseline:} A cat is laying on a bed and petting it. & \textbf{baseline:} A rider jumps over a fence.\\
 \textbf{capt:} a cat laying on a bed with a stuffed animal & \textbf{capt:} a horse is jumping over a wooden fence\\
 \textbf{VisCTG:} A cat laying on a bed being petted. & \textbf{VisCTG:} A rider jumps a fence on a horse.\\
  \hline
\end{tabular}}
\caption{\small Examples of retrieved images, associated captions, baseline and VisCTG (our visually grounded model's) generations for select concept sets. Note that the images and captions are used as an intermediary to guide the final generation 
and thus the final generation need not be faithful to them. E.g. there is nobody petting the cat in the image or caption, but since the VisCTG output is conditioned on both the concept set and the caption, it includes \textit{being petted}.}
\label{tab:image_caption_examples}
\end{table}

We investigate the task of generative commonsense reasoning, or CommonGen \cite{lin-etal-2020-commongen}, which involves generating sentences that effectively describe everyday scenarios from concepts sets, which are words that must appear in the output. CommonGen is challenging as effective relational reasoning ability using commonsense knowledge is required. Models must also possess the compositional generalization capabilities to piece together different concepts. CommonGen is an effective benchmark for constrained text generation and commonsense as its task formulation and evaluation methodology are rather broadly applicable.

We experiment on CommonGen using 
BART and T5. 
An initial analysis (\S\ref{sec:initial_qualitative_analysis}) of baseline generations shows several issues related to commonsense, specificity, and fluency. We hypothesize that these can be addressed through image captions (\S\ref{sec:initial_image_analysis}). Images representing everyday scenarios are commonplace, and typically logical and grounded in commonsense. Captioning models can also normally produce decent captions for everyday images, which can be used to guide and enhance the generation process. See Table \ref{tab:image_caption_examples} for examples.



Expounding on this, we use a pretrained image captioning model on MSCOCO captions \cite{lin2015microsoft} to caption the top retrieved images for each concept set (\S\ref{sec:image_retrieval_method},\ref{sec:image_captioning_method}). We use these captions as additional information to augment inputs to our generation models (\S\ref{sec:caption_selection_input_augmentation_method}). Extensive evaluation (\S\ref{sec:results_and_analysis}) demonstrates that VisCTG improves model performance and commonsense while addressing the baseline inadequacies.



\section{Dataset, Models, and Metrics}
\label{sec:dataset_models_metrics}
\subsection{CommonGen Dataset}
The original CommonGen dataset is made up of 35,141 concept sets (consisting of 3 to 5 keywords each) and 79,051 sentences, split into train, dev, and test splits. Since the original test set is hidden, we partition the original dev set into new dev and test splits for the majority of our experiments. We do, however, ask the CommonGen authors to evaluate our best VisCTG models on the original test set (more in \S\ref{sec:results_and_analysis}). The training set remains the same. We refer to the original dev and test sets as dev$_{O}$ and test$_{O}$, and these new splits as train$_{CG}$, dev$_{CG}$, and test$_{CG}$. Table \ref{tab:dataset_stats} contains information about these splits. Their relative sizes and distribution of concept set sizes within each are kept similar to the originals.

\begin{table}[t]
\centering
\small
\begin{tabular}{ |p{2.0cm}|p{1.0cm}|p{0.6cm}p{0.6cm}|p{0.8cm}p{0.8cm}| }
\hline
\textbf{Dataset Stats} & Train$_{CG}$ & Dev$_{O}$ & Test$_{O}$ & Dev$_{CG}$ & Test$_{CG}$ \\
\hline
\# concept sets & 32,651 & 993 & 1,497 & 240 & 360\\
  size = 3 & 25,020 & 493 & - & 120 & -\\
  size = 4 & 4,240 & 250 & 747 & 60 & 180\\
  size = 5 & 3,391 & 250 & 750 & 60 & 180\\
\hline
\end{tabular}
\caption{\small Statistics of CommonGen dataset splits.
}
\label{tab:dataset_stats}
\end{table}



\subsection{Models: T5 and BART}\label{sec:models_T5_BART}
We use pretrained text generation models T5 and BART, both the base and large versions. Both are seq2seq Transformer models. T5 has strong multitask pretraining. BART is pretrained as a denoising autoencoder to reproduce original from noised text. We use their HuggingFace implementations. 



We train two seeded versions of each model on train$_{CG}$ and evaluate their performance on dev$_{O}$. These serve as the baselines for our experiments. Using the numbers in \citet{lin-etal-2020-commongen} as comparison, we validate our implementations. We use the hyperparameters from \citet{lin-etal-2020-commongen}, beam search for decoding, and select the final epoch as the one reaching maximum ROUGE-2 \cite{lin2003automatic} on the dev split. From Table \ref{tab:reimplementation_stats}, we observe that our re-implementations reach or exceed reported results in \citet{lin-etal-2020-commongen} on most metrics.

\begin{table}[t]
\centering
\small
\begin{tabular}{|c|c|c|c|}
\hline
 \textbf{Model\textbackslash Metrics} &  \multicolumn{1}{c|}{BLEU-4} & CIDEr  & SPICE  \\
 \hline
 Reported BART-large    & 27.50 & 14.12  & 30.00   \\
 \hline
 Reported T5-base    & 18.00 & 9.73  & 23.40   \\
 \hline
 Reported T5-Large    & 30.60 & 15.84  & 31.80   \\
 \thickhline
 Our BART-base    & 28.30 & 15.07  & 30.35   \\
 \hline
 Our BART-large    & \textbf{30.20} & \textbf{15.72}  & \textbf{31.20}   \\ 
 \hline
 Our T5-base    & \textbf{31.00} & \textbf{16.37}  & \textbf{32.05}   \\
 \hline
 Our T5-large    & \textbf{33.60} & \textbf{17.02}  & \textbf{33.45}   \\
 \hline
\end{tabular}
\caption{\small Comparing dev$_{O}$ performance of our re-implemented models to those in \citet{lin-etal-2020-commongen}. Bold represents where we reach/exceed reported numbers. Results averaged over two seeds for our models. \citet{lin-etal-2020-commongen} did not report BART-base. See \S\ref{sec:eval_metrics} for metric explanations and Appendix \ref{appendix:reimplementation_numbers} for comparison of all metrics.}
\label{tab:reimplementation_stats}
\end{table}


\subsection{Evaluation Metrics}\label{sec:eval_metrics}
We use several evaluation metrics, including those in \citet{lin-etal-2020-commongen} such as 
BLEU \cite{papineni2002bleu}, 
CIDEr \cite{vedantam2015cider}, SPICE \cite{anderson2016spice}, and coverage (cov). These (other than cov) assess similarity between human references and generations. In particular, CIDEr captures a combination of sentence similarity, grammaticality, saliency, importance, and accuracy. SPICE maps texts to semantic scene graphs and calculates an F-score over these graphs' tuples. \citet{lin-etal-2020-commongen} note that SPICE correlates highest with human judgment for CommonGen. Cov measures the average percentage of input concepts covered by the output text in any form.
 
We also use BERTScore \cite{zhang2019bertscore} and Perplexity (PPL). BERTScore measures BERT \cite{devlin-etal-2019-bert} embeddings similarity between individual tokens, serving as a more semantic rather than surface-level similarity measure. We multiply by 100 when reporting BERTScore. PPL serves as a measure of fluency, with lower values representing higher fluency. 
We use GPT-2 \cite{radford2019language} for PPL. 
For all metrics other than PPL, higher means better performance.

\section{Initial Analysis and Motivation}
\label{sec:initial_analysis}
\subsection{Baseline Model Generations}\label{sec:initial_qualitative_analysis}
We conduct an initial analysis of the baseline model outputs, and observe that several lack fluency. Some are more like phrases than full coherent sentences, e.g. \textit{``body of water on a raft"}. Others miss important words, e.g. \textit{``A listening music and dancing in a dark room"} misses a noun before \textit{listening}. A large portion of generations are generic and bland, e.g. \textit{``Someone sits and listens to someone talk"}. 
This may be an instance of the \textit{dull response problem} faced by generation models \cite{du-black-2019-boosting,li2015diversity}, where they prefer safe and frequent responses independent of the input.

Many generations also lack commonsense. For example, \textit{``body of water on a raft"} is illogical as the phrases \textit{``body of water"} and \textit{``a raft"} are pieced together incorrectly. 
A similar issue occurs with the \textit{\{horse, carriage, draw\}} example in Table \ref{tab:qualitative_baselines}. 
At times the models also cannot understand what certain nouns can do, e.g. \textit{``A dog checking his phone on a pier."} Several other examples of this can be found in Table \ref{tab:qualitative_baselines}.

\begin{table*}[t]
\centering
\small
\begin{tabular}{ |c|c|c| } 
 \hline
 \textbf{Concept Set} & \textbf{Baseline Generation} & \textbf{Human Reference} \\
 \hline
 \{horse, carriage, draw\} & horse drawn in a carriage & The carriage is drawn by the horse.\\
 \hline
 \{dog, house, eat\} & A dog eats hay in a house & The dog eats food inside the house.\\
  \hline
 \{cow, horse, lasso\} & A cow is lassoing a horse. & A group of men riding horses lassoing a cow.\\
 \hline
\end{tabular}
\caption{\small Example generations from our baseline models versus human references.}
\label{tab:qualitative_baselines}
\end{table*}


\subsection{Images and Captions}\label{sec:initial_image_analysis}
Images that represent everyday scenarios are quite prevalent for almost any reasonable concept set. Further, the images are typically grounded in commonsense. For example, searching \textit{\{cow, horse, lasso\}} will result in many images of cowboys riding horses and lassoing cows, rather than the illogical situation of \textit{``A cow is lassoing a horse."} described by the baseline generation in Table \ref{tab:qualitative_baselines}. Many everyday images are relatively similar to those in image captioning datasets such as MSCOCO, so pretrained captioning models should work quite effectively. We thus hypothesize that using images and their captions to visually ground concept-to-text generation can potentially deal with issues mentioned in \ref{sec:initial_qualitative_analysis}. Retrieved images with corresponding captions generated by a pretrained image captioning model (see \S\ref{sec:image_captioning_method}) and final baseline and VisCTG generations for select concept sets are in Table \ref{tab:image_caption_examples}.

Textual corpora also suffer from \emph{reporting bias} \cite{gordon2013reporting}, where everyday, commonsense albeit ``uninteresting" actions (walking), objects (bench) and facts (bananas are yellow) are underrepresented compared to real-world frequency, while ``newsworthy" actions (murdering), objects (spaceships) and facts (blue GMO bananas) are exaggerated. This seeps into large pretrained text models \cite{shwartz2020neural}. Using visual data and models dampens this bias, likely improving the commonsense of generations. 

\section{Methodology}
\label{sec:methodology}
\begin{figure}
\centering
\includegraphics[width=0.47\textwidth]{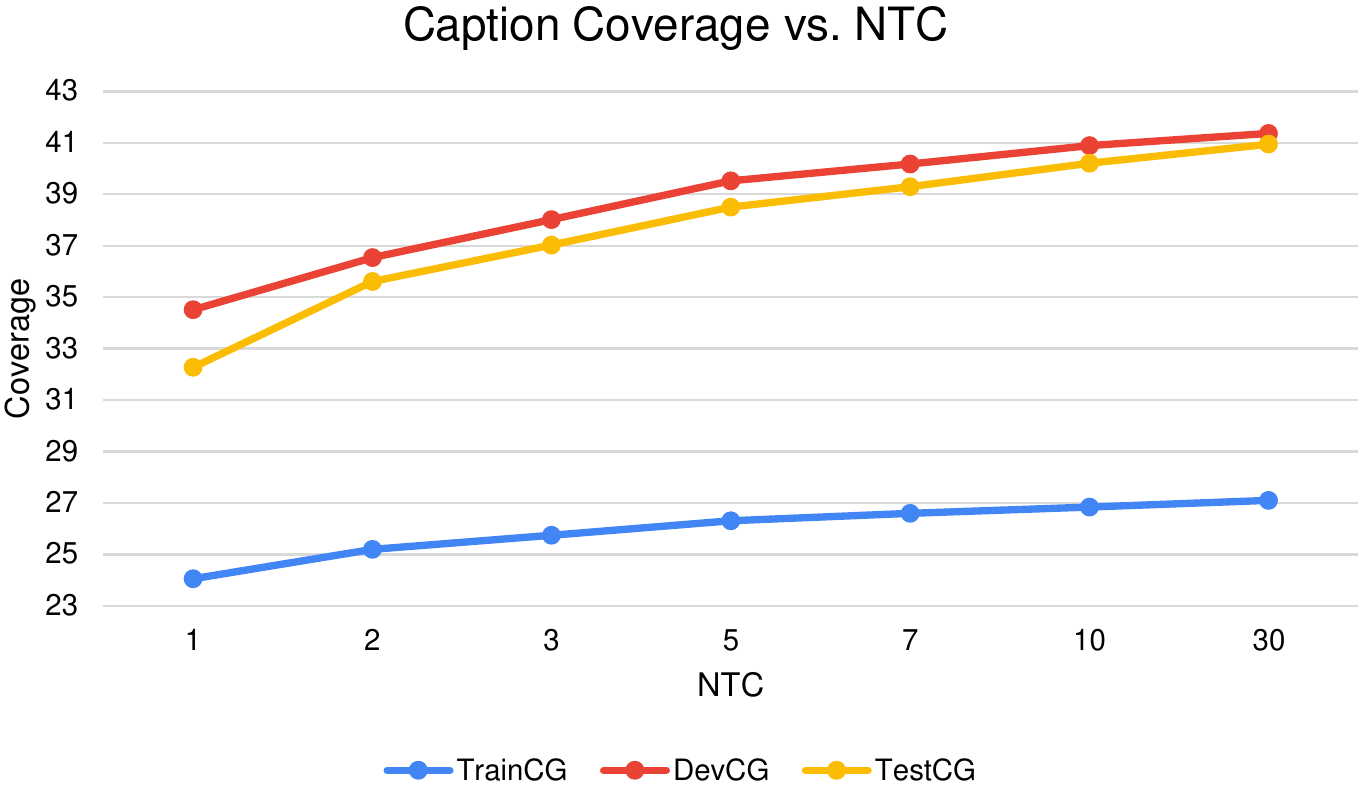}
\caption{\small Graph displaying the average coverage (out of 100) by the top NTC captions in aggregate per concept set.}
\label{fig:coverage_graph}
\end{figure}

\begin{table*}[t]
\centering
\small
\begin{tabular}{ |p{16.2cm}| } 
 \hline
 \begin{tightcenter}\vspace{-1.5mm}\textbf{Augmented Input} $\rightarrow$ \textbf{Final Generation}\vspace{-2.5mm}\end{tightcenter} \\
 \hline
 {wave fall board surfer} {$<$s$>$} {a surfer riding a wave on a surfboard} $\rightarrow$ \textbf{A surfer is falling off his board into the waves.}\\
 \hline
 {dance stage front crowd} {$<$s$>$} {a crowd of people watching a man on a stage} {$<$s$>$ } {a man is holding a microphone in front of a crowd} $\rightarrow$  \textbf{A man dances in front of a crowd on stage.}\\
 \hline
 {stand hold umbrella street} {$<$s$>$} {a woman walking down a street holding an umbrella} {$<$s$>$} {a woman walking down a street holding an umbrella} {$<$s$>$} {a girl holding a pink umbrella in a city} {$<$s$>$} {a man holding an umbrella in a city} {$<$s$>$} {a group of people standing under a umbrella} $\rightarrow$  \textbf{A group of people standing on a street holding umbrellas.}\\[-0.1ex]
 \hline
\end{tabular}
\caption{\small Examples of augmented inputs and final generations for varying values of NTC.}
\label{tab:augmented_concept_sets}
\end{table*}

\subsection{Image Retrieval}\label{sec:image_retrieval_method}
We first obtain images for each concept set in our three splits. Image captioning datasets such as MSCOCO and Flickr are typically too small and focused to be effective for our purposes since we must cover numerous different concept sets. Further, a search engine is more generalizable.

We decide to use Google Images. 
On a sample of concept sets, the retrieved images using other search engines were inappropriate; they did not incorporate most input keywords nor handle 
homonyms well. For example, 
\textit{``sports+fan+watch"} yields images of fans watching a sports game on Google images, but images of hand watches on Bing and DuckDuckGo.

We queried input concept sets by concatenating keywords with plus signs (+), and used \textit{simple-image-scraper}\footnote{\url{https://pypi.org/project/simple-image-download/}} to obtain URLs of the top 30 results. The image was scraped only if the URL ended in \textit{.png}, \textit{.jpeg}, \textit{.jpg}, or \textit{.gif}. The received content was verified to be valid images using \textit{pillow}\footnote{\url{https://pypi.org/project/Pillow/}}, otherwise skipped. Retrieved images were typically of high quality and corresponded well to the concepts. See Table \ref{tab:image_caption_examples} for examples.

\subsection{Image Captioning}\label{sec:image_captioning_method}
After retrieving images, we use a PyTorch-based implementation\footnote{\url{https://github.com/ruotianluo/self-critical.pytorch}} of 
the FC image captioning model \cite{luo2018discriminability,Rennie_2017_CVPR}, which generates a caption via an LSTM initialized with a pseudo token obtained by feeding the image into a deep CNN followed by a linear projection. 
We use a pretrained FC model trained on the MSCOCO dataset with pretrained Resnet-101 image features.\footnote{See Appendix \ref{appendix:image_captioning_model} for further captioning model details.} As most of our retrieved images represent everyday scenarios and are relatively similar to those in MSCOCO, the pretrained model performs quite well. See example captions in Table \ref{tab:image_caption_examples}.


\subsection{Caption Selection and Input Augmentation}\label{sec:caption_selection_input_augmentation_method}

After we have captions $S_{c} = \{c_{1},c_{2},...,c_{n}\}$ for each concept set in all three splits, 
we reorder them by descending coverage to the concept set to obtain 
$S_{c'} = \{c'_{1},c'_{2},...,c'_{n}\}$. If two captions are tied for coverage, we keep them in their original search result order. This allows us to select the captions that have highest coverage and are most relevant.

Since most retrieved images and corresponding captions cover only a fraction of the entire concept set, and the quality of each varies, we hypothesize that using multiple captions for generation may lead to more robust and higher-quality outputs with more coverage. The models may learn to piece together information from caption(s) while generating final texts. Hence, we try experiments using different numbers of top captions within $S_{c'}$, a parameter we call $NTC$ (Number of Top Captions). We try $NTC = 1,2,3,5,7,10$, and do not go above $NTC = 10$ as Figure \ref{fig:coverage_graph} shows that coverage gains from 10 $\rightarrow$ 30 are minor. Figure \ref{fig:coverage_graph} also illustrates that captions have relatively low individual coverage, especially compared with outputs from models trained on CommonGen, 
which is why we do not use them as a baseline.

The captions are concatenated together and onto the concept set using $<$s$>$ separator tokens. 
These serve as augmented inputs to BART and T5. 
They learn to convert these augmented inputs to human references during training, and are fed the augmented inputs (corresponding to the value of NTC) during validation and testing. 
Some examples of augmented inputs and generations can be found in Table \ref{tab:augmented_concept_sets}.

\section{Experiments}
\label{sec:experiments}
\subsection{Model Training and Selection}
For training VisCTG models, we mainly follow baseline hyperparameters, barring learning rate (LR) which is tuned per NTC value, and the maximum encoder length which is chosen depending on the 
tokenizer and value of NTC to ensure the entire input sequence can fit onto the encoder. 
We train two seeds per model. See Appendix \ref{appendix:model_training_finetuning_details} for more details.

For each model, we choose the epoch corresponding to highest ROUGE-2 on dev$_{CG}$, and use beam search for decoding. NTC itself is a hyperparameter, so while we train separate versions of each model corresponding to different NTC values, the final chosen models correspond to the NTC values that performed best on dev$_{CG}$ when averaged over both seeds. We then use the final chosen models to generate on both test$_{CG}$ and test$_{O}$, and report the results in \S\ref{sec:results_and_analysis}.

\subsection{Human Evaluation}\label{sec:human_eval}
We conduct two human evaluations: one using Amazon Mechanical Turk (AMT), and one using an expert linguist.\footnote{See Appendix \S\ref{sec:appendix_human_eval} for further human evaluation details.} For the AMT study, we ask annotators to evaluate 86 test$_{CG}$ examples per model. Our evaluation is based on pairwise comparison of VisCTG and baseline model outputs. We ask human annotators to choose which amongst the two outputs (presented in a random order per example) has better \textit{Overall Quality}. There are 3 choices - O1: VisCTG is better, O2: baseline is better, O3: both are indistinguishable. To aggregate multiple annotations per example, we find the fraction of responses towards each outcome value as the per-example distribution. We then find the sample mean of this outcome distribution over all examples. For sample mean and significance testing, we are interested in the values for O1 vs. O2.

For the expert linguist study, our expert is a native English speaker with a graduate degree in linguistics from a North American university. The expert is asked to annotate three aspects for 50 BART-large\footnote{Since this is the best performing VisCTG model 
- see \S\ref{sec:results_and_analysis}.} test$_{CG}$ examples - \textit{Overall Quality (Overall)}, \textit{Commonsense Plausibility (Commonsense)}, and \textit{Fluency (Fluency)}. For all aspects, we have a pairwise-comparison evaluation setup 
similar to that for AMT.

\section{Results and Analysis}
\label{sec:results_and_analysis}
\begin{table*}[t]
\centering
\small
\begin{tabular}{ |c|c|c|c|c|c|c| }
\hline
 & \multicolumn{3}{c|}{\textbf{BART-base} ($NTC=5$)} & \multicolumn{3}{c|}{\textbf{BART-large} ($NTC=2$)}\\
 \hline
 \underline{\textbf{Metrics}} & Baseline & VisCTG & p-value & Baseline & VisCTG & p-value\\
 \hline
 ROUGE-1 & 43.96$\pm0.03$ & \textbf{45.44}$\pm$0.08 & 1.58E-05 & 45.67$\pm0.25$ & \textbf{46.91}$\pm$0.31 & 1.58E-05\\
 \hline 
 ROUGE-2 & 17.31$\pm0.02$ & \textbf{19.15}$\pm$0.21 & 1.58E-05 & 18.77$\pm0.04$ & \textbf{20.36}$\pm$0.05 & 1.58E-05\\
 \hline
 ROUGE-L & 36.65$\pm0.00$ & \textbf{38.43}$\pm$0.07 & 1.58E-05 & 37.83$\pm0.29$ & \textbf{39.23}$\pm$0.01 & 1.58E-05\\
 \hline
 BLEU-1 & 73.20$\pm0.28$ & \textbf{75.65}$\pm$0.78 & 6.94E-05 & 74.45$\pm0.21$ & \textbf{78.80}$\pm$0.28 & 6.94E-05\\
 \hline
 BLEU-2 & 54.50$\pm0.14$ & \textbf{59.05}$\pm$0.07 & 6.94E-05 & 56.25$\pm0.78$ & \textbf{61.60}$\pm$0.85 & 6.94E-05\\
 \hline
 BLEU-3 & 40.40$\pm0.14$ & \textbf{44.90}$\pm$0.42 & 6.94E-05 & 42.15$\pm0.49$ & \textbf{47.00}$\pm$0.71 & 6.94E-05\\
 \hline
 BLEU-4 & 30.10$\pm0.14$ & \textbf{34.10}$\pm$0.57 & 3.82E-03 & 32.10$\pm0.42$ & \textbf{36.25}$\pm$0.78 & 2.08E-04\\
 \hline
 METEOR & 30.35$\pm0.35$ & \textbf{31.95}$\pm$0.07 & 6.94E-05 & 31.70$\pm0.14$ & \textbf{34.00}$\pm$0.14 & 6.94E-05\\
 \hline
 CIDEr & 15.56$\pm0.10$ & \textbf{16.84}$\pm$0.05 & 6.94E-05 & 16.42$\pm0.09$ & \textbf{18.35}$\pm$0.13 & 6.94E-05\\
 \hline
 SPICE & 30.05$\pm0.07$ & \textbf{31.80}$\pm$0.28 & 6.94E-05 & 31.85$\pm0.21$ & \textbf{34.60}$\pm$0.28 & 6.94E-05\\
 \hline
 BERTScore & 59.19$\pm0.32$ & \textbf{61.44}$\pm$0.02 & 1.58E-05 & 59.95$\pm0.29$ & \textbf{62.85}$\pm$0.30 & 1.58E-05\\
 \hline
 Coverage & 90.43$\pm0.17$ & \textbf{90.66}$\pm$1.39 & 0.33* & 94.49$\pm0.53$ & \textbf{96.49}$\pm$0.24 & 1.58E-05\\
 \hline
 PPL & 80.39$\pm3.65$ & \textbf{72.45}$\pm$0.79 & 1.58E-05 & 80.37$\pm$4.51 & \textbf{68.46}$\pm$5.90 & 1.58E-05\\
 \hline
\end{tabular}
\caption{\small Automatic eval results for BART on test$_{CG}$ over two seeds. Bold corresponds to best performance on that metric. 
We include stat sig p-values (from Pitman's permutation test \cite{pitman1937significance}) for VisCTG compared to the baseline. Insignificant ones ($\alpha=0.1$) marked with *.}
\label{tab:automatic_results_BART}
\end{table*}

\begin{table*}[t]
\centering
\small
\begin{tabular}{ |c|c|c|c|c|c|c| }
\hline
 & \multicolumn{3}{c|}{\textbf{T5-base} ($NTC=2$)} & \multicolumn{3}{c|}{\textbf{T5-large} ($NTC=1$)}\\
 \hline
 \underline{\textbf{Metrics}} & Baseline & VisCTG & p-values & Baseline & VisCTG & p-values\\
 \hline
 ROUGE-1 & 44.63$\pm0.13$ & \textbf{46.26}$\pm$0.07 & 1.58E-05 & 46.32$\pm$0.26 & \textbf{46.93}$\pm$0.22 & 7.26E-04\\
 \hline 
 ROUGE-2 & 18.40$\pm0.14$ & \textbf{19.78}$\pm$0.30 & 1.58E-05 & 19.59$\pm$0.12 & \textbf{20.01}$\pm$0.23 & 0.02\\
 \hline
 ROUGE-L & 37.60$\pm0.16$ & \textbf{38.91}$\pm$0.27 & 1.58E-05 & 39.20$\pm$0.21 & \textbf{39.52}$\pm$0.43 & 0.06\\
 \hline
 BLEU-1 & 73.60$\pm0.85$ & \textbf{76.80}$\pm$0.28 & 6.94E-05 & 77.55$\pm$0.35 & \textbf{78.65}$\pm$0.21 & 4.65E-03\\
 \hline
 BLEU-2 & 57.00$\pm0.71$ & \textbf{60.30}$\pm$0.28 & 6.94E-05 & 60.80$\pm$0.28 & \textbf{61.55}$\pm$0.35 & 0.07\\
 \hline
 BLEU-3 & 42.75$\pm0.49$ & \textbf{46.25}$\pm$0.64 & 6.94E-05 & 46.50$\pm$0.00 & \textbf{47.10}$\pm$0.57 & 0.11*\\
 \hline
 BLEU-4 & 32.70$\pm0.42$ & \textbf{36.10}$\pm$0.85 & 6.94E-05 & 36.20$\pm$0.14 & \textbf{36.40}$\pm$0.28 & 0.21*\\
 \hline
 METEOR & 31.05$\pm0.49$ & \textbf{32.70}$\pm$0.00 & 6.94E-05 & 33.20$\pm$0.00 & \textbf{33.65}$\pm$0.49 & 0.49*\\
 \hline
 CIDEr & 16.26$\pm0.25$ & \textbf{17.65}$\pm$0.02 & 6.94E-05 & 17.79$\pm$0.01 & \textbf{17.94}$\pm$0.25 & 0.23*\\
 \hline
 SPICE & 31.95$\pm0.07$ & \textbf{33.40}$\pm$0.28 & 6.94E-05 & 33.90$\pm$0.42 & \textbf{34.55}$\pm$0.21 & 0.03\\
 \hline
 BERTScore & 61.40$\pm0.34$ & \textbf{62.42}$\pm$0.17 & 1.58E-05 & 62.67$\pm$0.09 & \textbf{62.72}$\pm$0.03 & 0.34*\\
 \hline
 Coverage & 90.96$\pm1.77$ & \textbf{94.48}$\pm$1.39 & 1.58E-05 & 94.40$\pm$0.02 & \textbf{95.95}$\pm$0.45 & 1.58E-05\\
 \hline
 PPL & 83.04$\pm$1.62 & \textbf{77.50}$\pm$3.86 & 3.16E-05 & 81.78$\pm$4.63 & \textbf{73.41}$\pm$4.32 & 1.58E-05\\
 \hline
\end{tabular}
\caption{\small Automatic eval results for T5 on test$_{CG}$ over two seeds. Bold corresponds to best performance on that metric. 
We include stat sig p-values (from Pitman's permutation test \cite{pitman1937significance}) for VisCTG compared to the baseline. Insignificant ones ($\alpha=0.1$) marked with *.}
\label{tab:automatic_results_T5}
\end{table*}

\begin{table*}[t]
\centering
\small
\begin{tabular}{ |c|cc|cc|c|cc|c| }
\hline
 \textbf{Models\textbackslash Metrics} & \multicolumn{2}{c|}{ROUGE-2/L} & \multicolumn{2}{c|}{\textbf{BLEU}-3/\textbf{4}} & METEOR & \textbf{CIDEr} & \textbf{SPICE} & Coverage\\
 \hline
 T5-base (reported baseline) & 14.63 & 34.56 & 28.76 & 18.54 & 23.94 & 9.40 & 19.87 & 76.67 \\
 \hline
 T5-large (reported baseline) & 21.74 & 42.75 & 43.01 & 31.96 & 31.12 & 15.13 & 28.86 & 95.29 \\
\hline
 BART-large (reported baseline) & 22.02 & 41.78 & 39.52 & 29.01 & 31.83 & 13.98 & 28.00 & 97.35 \\
 \Xhline{2\arrayrulewidth}
 EKI-BART \cite{fan2020enhanced} & - & - & - & 35.945 & - & 16.999 & 29.583 & - \\
 \hline
 KG-BART \cite{liu2020kg} & - & - & - & 33.867 & - & 16.927 & 29.634 & - \\
 \hline
 SAPPHIRE (T5-large) \cite{feng-etal-2021-sapphire} & - & - & - & 37.119 & - & 16.901 & 29.751  & - \\
 \hline
 RE-T5 \cite{wang2021retrieval} & - & - & - & \textbf{40.863} & - & \textbf{17.663} & \textbf{31.079} & - \\
 \Xhline{2\arrayrulewidth}
 T5-base VisCTG & 22.83 & 44.98 & 45.749 & \textbf{34.722} & 31.809 & \textbf{16.173} & \textbf{28.808} & 92.92 \\ 
 \hline
 T5-large VisCTG & 23.83 & 45.76 & 47.376 & 36.409 & 33.012 & 16.815 & 29.629 & 95.54 \\ 
 \hline 
 BART-base VisCTG & 21.73 & 43.43 & 43.235 & \textbf{32.291} & 30.86 & \textbf{15.187} & \textbf{27.403} & 88.98 \\ 
 \hline
 BART-large VisCTG & 23.68 & 45.07 & 48.031 & \textbf{36.939} & 33.215 & \textbf{17.199} & \textbf{29.973} & 94.86 \\ 
 \hline
\end{tabular}
\caption{\small Automatic eval results of VisCTG models on test$_{O}$, evaluated by CommonGen authors. We compare to reported baseline numbers in \citet{lin-etal-2020-commongen} (they did not evaluate BART-base), and models on their leaderboard\footnoteref{leaderboard} with publications at time of writing that outperform baselines. Their leaderboard reports BLEU-4, CIDEr, and SPICE. Bold corresponds to best performance (for those three) per model type+size.}
\label{tab:results_original_test_split}
\end{table*}

\begin{table}[!ht]
\centering
\small
\begin{tabular}{ |c|c|c|c||c| }
\hline
\underline{Model} & \textbf{O1} & \textbf{O2} & \textbf{O3} & \textbf{IAA}\\
 \hline
 \textbf{BART-base} & \textbf{0.45} & 0.33 & 0.22 & 0.72\\
 \hline
 \textbf{BART-large} & \textbf{0.62} & 0.18 & 0.20 & 0.55\\
 \hline
 \textbf{T5-base} & \textbf{0.46} & 0.33 & 0.21 & 0.72\\
 \hline
 \textbf{T5-large} & \textbf{0.46} & 0.34 & 0.20 & 0.74\\
 \hline
\end{tabular}
\captionof{table}{\small Avg. AMT eval results on test$_{CG}$ for \textit{overall quality}. O1: VisCTG wins, O2: baseline wins, O3: both indistinguishable. Bold corresponds to higher fractional outcome between O1 and O2. All results are statistically significant based on paired two-tailed t-tests and $\alpha=0.1$. The inter-annotator agreement (IAA) is the average direct fractional agreement (where both annotators choose O1 or O2) over all examples. 
See \S\ref{sec:human_eval} and Appendix \ref{sec:appendix_human_eval} for further details.}
\label{tab:human_eval_results}
\end{table}

\begin{table}[!ht]
\centering
\small
\begin{tabular}{ |c|c|c|c|c| }
\hline
\underline{Model} & \underline{Aspect} & \textbf{O1} & \textbf{O2} & \textbf{O3}\\
 \hline
 \multirow{3}{*}{\textbf{BART-large}} & Overall & \textbf{0.44} & 0.24 & 0.32 \\
 \cline{2-5}
 & Commonsense & \textbf{0.32} & 0 & 0.68 \\
 \cline{2-5}
 & Fluency & \textbf{0.56} & 0.12 & 0.32 \\
 \hline
\end{tabular}
\captionof{table}{\small Avg. expert linguist eval results on test$_{CG}$ for BART-large. O1: VisCTG wins, O2: baseline wins, O3: both indistinguishable. Bold corresponds to higher fractional outcome between O1 and O2 per aspect. See \S\ref{sec:human_eval} and Appendix \ref{sec:appendix_human_eval} for further details.}
\label{tab:human_eval_results_linguist}
\end{table}

\begin{table*}[ht!]
\centering
\small
\addtolength{\tabcolsep}{-4pt}
\begin{tabular}{|p{2.7cm}|p{14cm}|}
\hline 
\textbf{Method} & \textbf{Text} \\ \hline
Concept set & \{sit, chair, toy, hand\} (example 1)\\ \hline
Captions & {a little girl sitting on a chair with a teddy bear} $<$s$>$ {a small child sitting on a chair with a teddy bear} $<$s$>$ {a young boy sitting on a chair with a skateboard} $<$s$>$ 
{a man sitting on a chair with a remote} \\ \hline
BART-base-BL & {hands sitting on a chair} \\ \hline
BART-base-VisCTG & {A boy sitting on a chair with a toy in his hand.} \\ \hline 
Human reference & {A baby sits on a chair with a toy in one of its hands.} \\
\Xhline{3\arrayrulewidth}
Concept set & \{food, eat, hand, bird\} (example 2)\\ \hline
Captions & {a bird is perched on a branch with a hand} $<$s$>$ {a person holding a small bird in their hand}\\ \hline
BART-large-BL & {hand of a bird eating food} \\ \hline
BART-large-VisCTG & {A bird eats food from a hand.} \\ \hline 
Human reference & {A small bird eats food from someone's hand.} \\
\Xhline{3\arrayrulewidth}
Concept set & \{bench, bus, wait, sit\} (example 3)\\ \hline
Captions & {a man sitting on a bench with a book} $<$s$>$ {a person sitting on a bench with a laptop}\\ \hline
T5-base-BL & {A bus sits on a bench.} \\ \hline
T5-base-VisCTG & {A man sits on a bench waiting for a bus.} \\\hline 
Human reference & {The man sat on the bench waiting for the bus.} \\
\Xhline{3\arrayrulewidth}
Concept set & \{jacket, wear, snow, walk\} (example 4)\\ \hline
Captions & {a young boy in a red jacket is standing in the snow} $<$s$>$ {a man in a red jacket is standing in the snow}\\ \hline
BART-large-BL & {walking in the snow wearing a furry jacket} \\ \hline
BART-large-VisCTG & {A man is walking in the snow wearing a jacket.} \\ \hline 
Human reference & 
{Jamie took a walk out into the snow with only a T shirt on and instantly went back inside to wear his jacket.}
\\\Xhline{3\arrayrulewidth}
Concept set & \{hold, hand, stand, front\} (example 5)\\ \hline
Captions & {a man holding a pair of scissors in front of a wall}\\ \hline
T5-large-BL & {Someone stands in front of someone holding a hand.} \\ \hline
T5-large-VisCTG & {A man stands in front of a man holding a hand.} \\\hline 
Human reference & {A man stands and holds his hands out in front of him.} \\
\Xhline{3\arrayrulewidth}
Concept set & \{bag, put, apple, tree, pick\} (example 6)\\ \hline
Captions & {a person holding a apple in a tree} $<$s$>$ {a bunch of apples are growing on a tree} $<$s$>$ {a close up of a green apple with a tree} $<$s$>$ {a bunch of apples are growing on a tree} 
\\ \hline
BART-base-BL & {A man is putting apples in a bag and picking them up from the tree.} \\ \hline
BART-base-VisCTG & {A man puts a bag of apples on a tree.} \\ \hline 
Human reference & {I picked an apple from the tree and put it in my bag.} \\
\hline \end{tabular}
\caption{\small Qualitative examples for test$_{CG}$. \textit{BL} stands for baseline. \textit{Concept set} refers to the input keywords and \textit{Captions} refers to the {captions} (separated by $<$s$>$) used by the VisCTG model for that particular example to produce its final generation.}
\label{tab:qualitative_body}
\end{table*}

\begin{figure*}[t]
    \centering
    \includegraphics[width=0.32\textwidth]{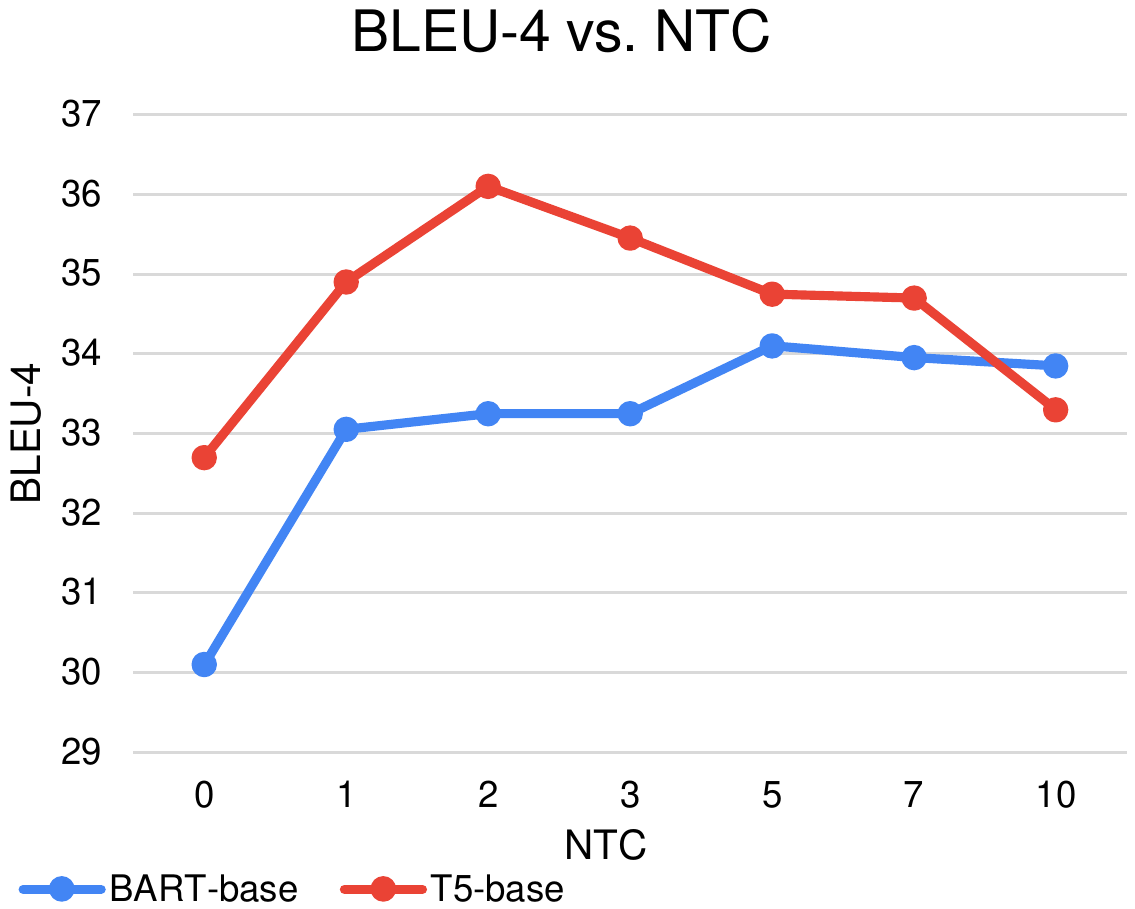}
    \includegraphics[width=0.32\textwidth]{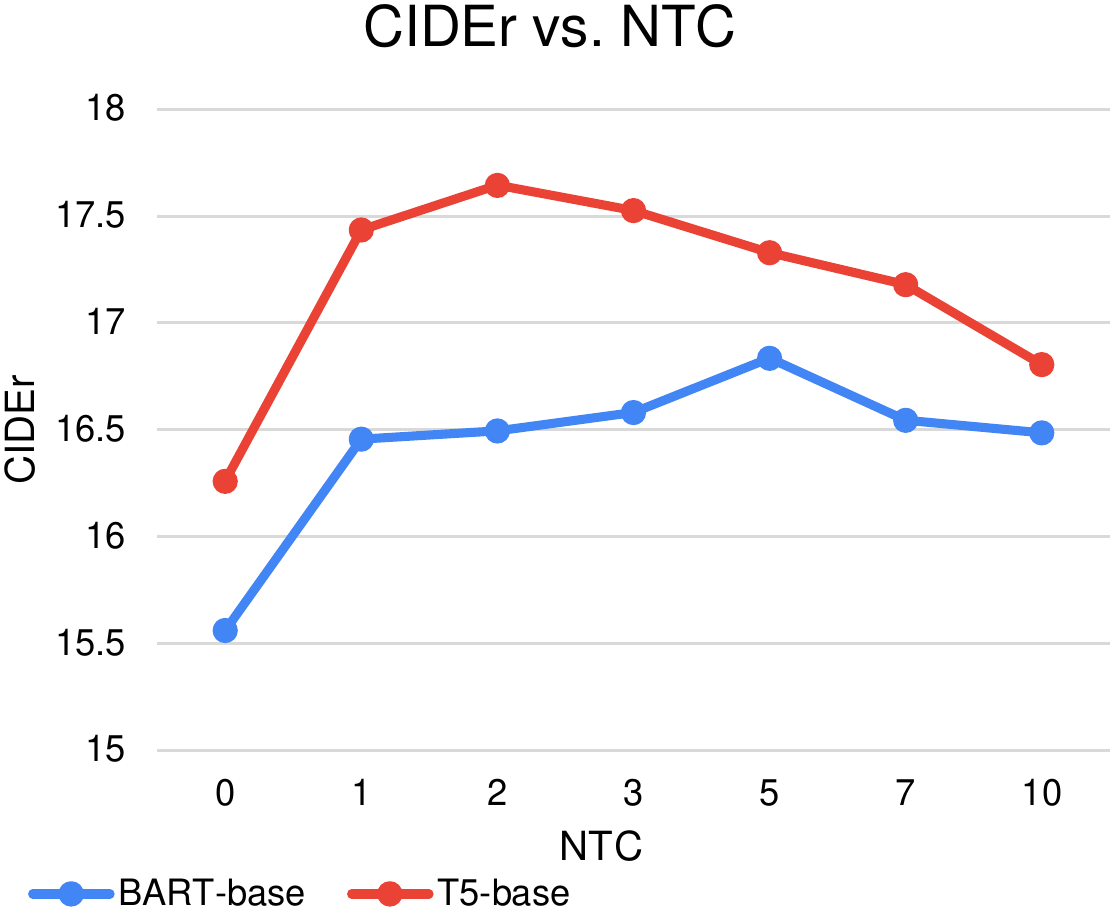}
    \includegraphics[width=0.32\textwidth]{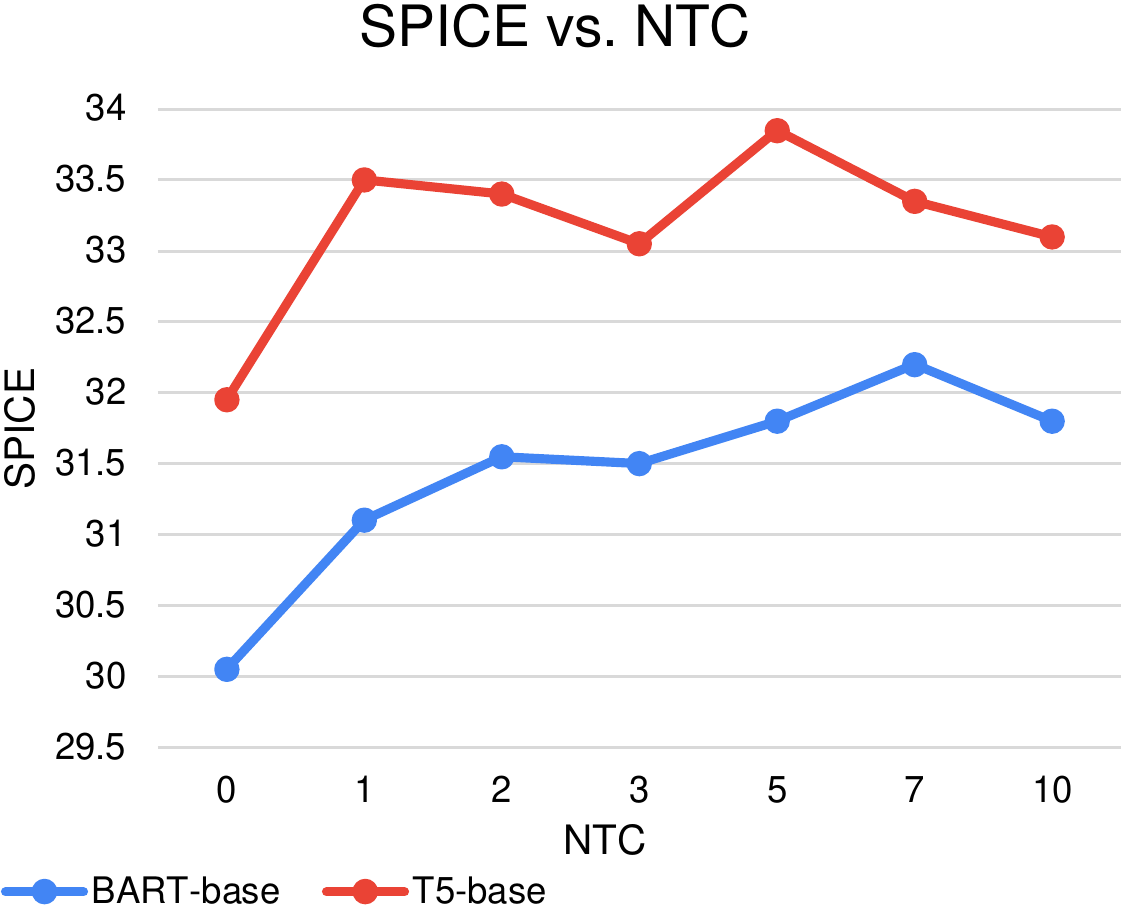}
    \caption{\small BLEU-4, CIDEr, and SPICE on test$_{CG}$ over different values of NTC for BART-base and T5-base.} \label{fig:NTC_graphs}
\end{figure*}

Automatic evaluation results on test$_{CG}$ are in Tables \ref{tab:automatic_results_BART} and \ref{tab:automatic_results_T5}, and results on test$_{O}$ in Table \ref{tab:results_original_test_split}.
\footnote{Evaluated by the CommonGen authors on their hidden test set.} Graphs displaying BLEU-4, CIDEr, and SPICE (the metrics on the CommonGen leaderboard\footnote{\label{leaderboard}\url{https://inklab.usc.edu/CommonGen/leaderboard.html}}) on test$_{CG}$ over different NTC values are in Figure \ref{fig:NTC_graphs}. Human evaluation results on test$_{CG}$ are in Tables~\ref{tab:human_eval_results} and ~\ref{tab:human_eval_results_linguist}. Optimal NTC values for BART-base, BART-large, T5-base, and T5-large are 5, 2, 2, and 1, respectively. These are the VisCTG results reported in the aforementioned tables. Table \ref{tab:qualitative_body} contains qualitative examples, with more in Appendix \S\ref{sec:appendix_qualitative_examples}.

\subsection{Analysis of Automatic Evaluation Results}\label{sec:automatic_eval}
We see from Tables \ref{tab:automatic_results_BART} and \ref{tab:automatic_results_T5} that VisCTG outperforms the baselines on all metrics across the models on test$_{CG}$. Performance gains are strong and statistically significant for BART-base, BART-large, and T5-base. VisCTG appears relatively less effective for T5-large which is the strongest baseline, and hence improving its performance may be more difficult.

From Table \ref{tab:results_original_test_split}, we see that VisCTG models substantially outperform corresponding baselines reported in \citet{lin-etal-2020-commongen} on test$_O$. T5-base VisCTG outperforms the reported T5-base and large baselines across metrics, and BART-base VisCTG performs similarly to the reported BART-large baseline. BART-large VisCTG outperforms the reported baseline, EKI-BART \cite{fan2020enhanced}, and KG-BART \cite{liu2020kg}. These are SOTA published CommonGen BART models that use external knowledge from corpora and KGs. We show that visual grounding is more effective, and BART-large VisCTG places high on the leaderboard.\footnoteref{leaderboard} T5-large VisCTG outperforms the reported baseline, but lags behind SAPPHIRE \cite{feng-etal-2021-sapphire} and RE-T5 \cite{wang2021retrieval}.

Figure \ref{fig:NTC_graphs} shows that as NTC increases, BLEU-4, CIDEr, and SPICE increase to a peak, 
and taper off after. This is expected as we saw in Figure \ref{fig:coverage_graph} that the rate of increase of coverage declines with larger NTC. The latter images and captions are of diminishing quality, and hence using too many negatively affects model performance.

We also computed ROUGE between captions and outputs over test$_{CG}$. $ROUGE1/2/L=36.2/12.3/33.5$ are modestly valued. Our models do not simply copy caption content.

\subsection{Analysis of Human Evaluation Results}\label{sec:human_eval_analysis}
Table \ref{tab:human_eval_results} shows that VisCTG outperforms the baseline on all four models based on human annotators (with high IAA). 
Annotators, on average, prefer VisCTG outputs over baseline outputs on overall quality, especially for BART-large. Table \ref{tab:human_eval_results_linguist} illustrates that VisCTG outperforms the baseline model for BART-large based on an expert linguist's perspective. VisCTG outputs are highly preferred, on average, over the baseline on all three aspects of overall quality, commonsense, and fluency. This aligns with our automatic results in \S\ref{sec:automatic_eval}, where VisCTG outperforms the baselines across all models.

\subsection{Qualitative Analysis}\label{sec:post_qualitative_analysis}

Table \ref{tab:qualitative_body} shows several baseline outputs that contain issues from \S\ref{sec:initial_qualitative_analysis}, e.g. incomplete and/or illogical sentences. Human references are all fluent and logical. VisCTG can usually generate much higher-quality text than the baselines.

The baseline outputs for ex. 1-2 are phrases lacking arguments, and all illogical for ex. 1-3. Using captions, VisCTG successfully adjusts semantic roles of entities, replaces incorrect subjects, fixes dependency structure, and grounds generations in commonsense. For ex. 1, captions are of the form ``\{X\} \textit{sitting on a chair with} \{Y\}", where \{X\} is a subject and \{Y\} an object. VisCTG output has similar structure, being fluent and logical with higher coverage. The baseline output also has an incorrect subject of \textit{``hands"}. Our VisCTG output contains an additional entity (not present in the input set) of \textit{``boy"} as subject, likely since it is a subject in the captions. This highlights the usefulness of visual grounding, as the image space can provide additional commonsense information not present in the text (e.g. toys are associated with children/boys). For ex. 2, the baseline output treats \textit{``hand of a bird"} as a single entity, the subject. Captions separate \textit{``bird"} and \textit{``hand"} into two, likely guiding the VisCTG output to do so. For ex. 3, the baseline misplaces \textit{``bus"} as subject. Captions are of form ``\{X\} \textit{sitting on a bench} \{Y\}", where \{X\} is a logical subject and \{Y\} is an expression. The VisCTG output has this structure, with correct subject and commonsense, and higher coverage. Overall, we see that visual grounding guides the model to learn which nouns/subjects can perform which actions (e.g. \textit{``hands"} cannot sit on a chair but a \textit{``boy"} can), which is a major baseline deficiency discussed in \S\ref{sec:initial_qualitative_analysis}. 

For ex. 4, the baseline output lacks a subject that the captions contain, likely guiding the VisCTG output to contain one: \textit{``a man"}. For ex. 5, the baseline output is generic due to uses of \textit{``someone"}. VisCTG's output is more specific and refers to \textit{``man"}, likely because the caption (though not very fitting) includes a \textit{``man"} subject. Even for captions that fit the concepts less, structure and fluency can still be exploited.

Overall, we see that the baselines simply try to piece together the input concepts into a form of English syntax, often failing to do so effectively. 
VisCTG models can produce more grammatical, fluent, and logical text by exploiting the syntactic and dependency structures of the captions. Further, the visual grounding improves the commonsense of the generations. The images inherently capture commonsense by representing everyday scenarios, and this commonsense info is rarely explicitly included in text. Hence, large text-based models such as our baselines tend to not know this info, whereas VisCTG models learn it through the grounding.

VisCTG is, however, imperfect. For ex. 6, its output is less logical and lower coverage than the baseline's. The captions are all simplistic and low coverage; the first is illogical, and some others are of the form \textit{``a bunch of apples \{...\} on a tree"}, likely negatively impacting the generation. Ex. 4's human reference is creative, which is an area where VisCTG still lacks in comparison. For ex. 5, while VisCTG edits \textit{``someone"} to \textit{``man"}, it is unable to merge the two instances of \textit{``man"} or adjust the sentence to be more coherent. 
These weaknesses are likely because captions tend to be simplistic (due to the captioning model's training data), limiting VisCTG's ability to make heavier edits. VisCTG, unsurprisingly, appears to depend quite heavily on the captions, and hence the quality of the images and captioning model.

\section{Related Work}
\label{sec:related_work}
\noindent \textbf{Constrained Text Generation:} There have been several works on constrained text generation. 
\citet{miao2019cgmh} use Metropolis-Hastings sampling to determine Levenshtein edits per generation step. 
\citet{feng2019keep} propose Semantic Text Exchange to adjust topic-level text semantics. \citet{gangal2021nareor} introduce narrative reordering (NAREOR) to edit the temporality of narratives.


\paragraph{Data-to-text NLG:} E2E-NLG \cite{duvsek2018findings} and WebNLG \cite{gardent2017webnlg} are two popular NLG benchmarks with structured inputs - meaning representation (MR) and triple sequences, respectively. 
\citet{montella2020denoising} use Wiki sentences with parsed OpenIE triples as weak supervision for WebNLG. 

\paragraph{Commonsense Injection and Incorporation:}
One large commonsense knowledge graph (KG) is COMET, trained on KG edges to learn connections between words and phrases. 
EKI-BART \cite{fan2020enhanced} and KG-BART \cite{liu2020kg} use external knowledge (from corpora and KGs) to improve BART's performance on CommonGen. Distinctly, VisCTG uses visual grounding and shows higher performance (see \S\ref{sec:results_and_analysis}). SAPPHIRE \cite{feng-etal-2021-sapphire} consists of simple model-agnostic improvements on CommonGen that rely purely on the data itself and the model's initial generations. Visual Commonsense Reasoning (VCR) \cite{Zellers_2019_CVPR} involves answering commonsense-related multiple-choice questions about images. 
Our work uniquely focuses on injecting commonsense into seq2seq 
models 
for text generation.

\paragraph{Multimodal Machine Learning and NLP:} There has been more work on multimodality, 
in 
areas like 
representation 
and 
video captioning, 
but little for constrained and data-to-text NLG \cite{multimodal_survey_1,multimodal_survey_2}. There is work on pretrained multimodal models like ViLBERT \cite{NEURIPS2019_c74d97b0}, which are mainly encoders that jointly represent images and text rather than seq2seq models, and would be ill-suited for generation. Further, unlike these models which are pretrained, VisCTG exploits per-example visual information to fix specific issues for each concept set.



\section{Conclusion and Future Work}
\label{sec:conclusion}
In conclusion, we motivated and explored the use of visual grounding for improving the commonsense of Transformer models for text generation. We investigated this for concept-to-text generation, calling our method VisCTG: Visually Grounded Concept-to-Text Generation. Extensive experiments on BART and T5 showed its efficacy on the CommonGen task. Comprehensive evaluation and analysis showed that VisCTG boosts model performance and commonsense while addressing baseline deficiencies. Potential future work includes improving image search and captioning, such as better selection of images during retrieval or using a stronger captioning model, e.g. one based on CLIP \cite{https://doi.org/10.48550/arxiv.2103.00020}. 
Video captioning and image generation rather than retrieval can also be explored. Further, VisCTG can be investigated for other data-to-text NLG tasks, e.g. WebNLG, and applications like 
data augmentation for text generation \cite{feng-etal-2020-genaug,feng2021survey} and enhancing the commonsense reasoning of personalized dialogue agents \cite{Li_Jiang_Feng_Sprague_Zhou_Hoey_2020}. 

\bibliography{anthology,aaai22}

\clearpage
\appendix
\section*{Appendices}

\begin{table*}[ht]
\centering
\small
\begin{tabular}{|c|cc|cc|c|cc|c|c|}
\hline
 \textbf{Model\textbackslash Metrics} & \multicolumn{2}{c|}{ROUGE-2/L} & \multicolumn{2}{c|}{BLEU-3/4} & METEOR & CIDEr & SPICE & BERTScore & Cov\\
 \hline
 Reported BART-large & 22.13 & 43.02 & 37.00 & 27.50 & 31.00 & 14.12 & 30.00 & - & 97.56 \\
 \hline
 Reported T5-base & 15.33 & 36.20 & 28.10 & 18.00 & 24.60 & 9.73 & 23.40 & - & 83.77 \\
 \hline
 Reported T5-Large & 21.98 & 44.41 & 40.80 & 30.60 & 31.00 & 15.84 & 31.80 & - & 97.04 \\
 \thickhline
 Our BART-base & 15.91 & 36.15 & 38.30 & 28.30 & 30.20 & 15.07 & 30.35 & 58.26 & 93.44 \\
 \hline
 Our BART-large & 17.27 & 37.32 & \textbf{39.95} & \textbf{30.20} & \textbf{31.15} & \textbf{15.72} & \textbf{31.20} & 58.58 & 95.03 \\ 
 \hline
 Our T5-base & \textbf{17.27} & \textbf{37.69} & \textbf{41.15} & \textbf{31.00} & \textbf{31.10} & \textbf{16.37} & \textbf{32.05} & 60.32 & \textbf{94.44} \\
 \hline
 Our T5-large & 17.90 & 38.31 & \textbf{43.80} & \textbf{33.60} & \textbf{32.70} & \textbf{17.02} & \textbf{33.45} & 61.39 & 96.26 \\
 \hline
\end{tabular}
\caption{\small Performance of our re-implemented CommonGen models on dev$_{O}$ compared to the original numbers reported in \citet{lin-etal-2020-commongen}. Note that for our models, results are averaged over two seeds, and that the original authors did not experiment with BART-base or report BERTScore. Bold indicates where we match or exceed the corresponding reported baseline metric.}
\label{tab:full_reimplementation_stats}
\end{table*}

\begin{figure*}[ht]
\begin{subfigure}{.97\textwidth}
    \centering
    \includegraphics[width=0.99\textwidth]{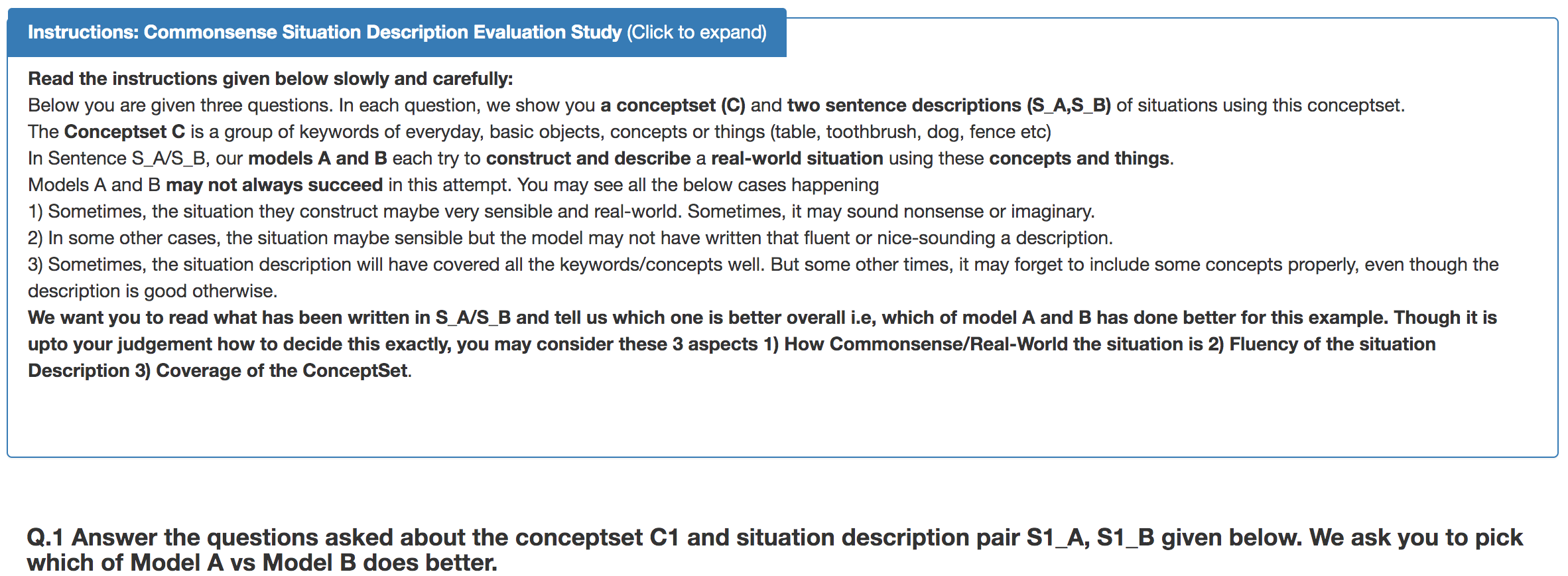}
    \caption{}
    \label{fig:amt_preserve_instructions_graph}
\end{subfigure} \\
\begin{subfigure}{.97\textwidth}
    \centering
    \includegraphics[width=0.99\textwidth]{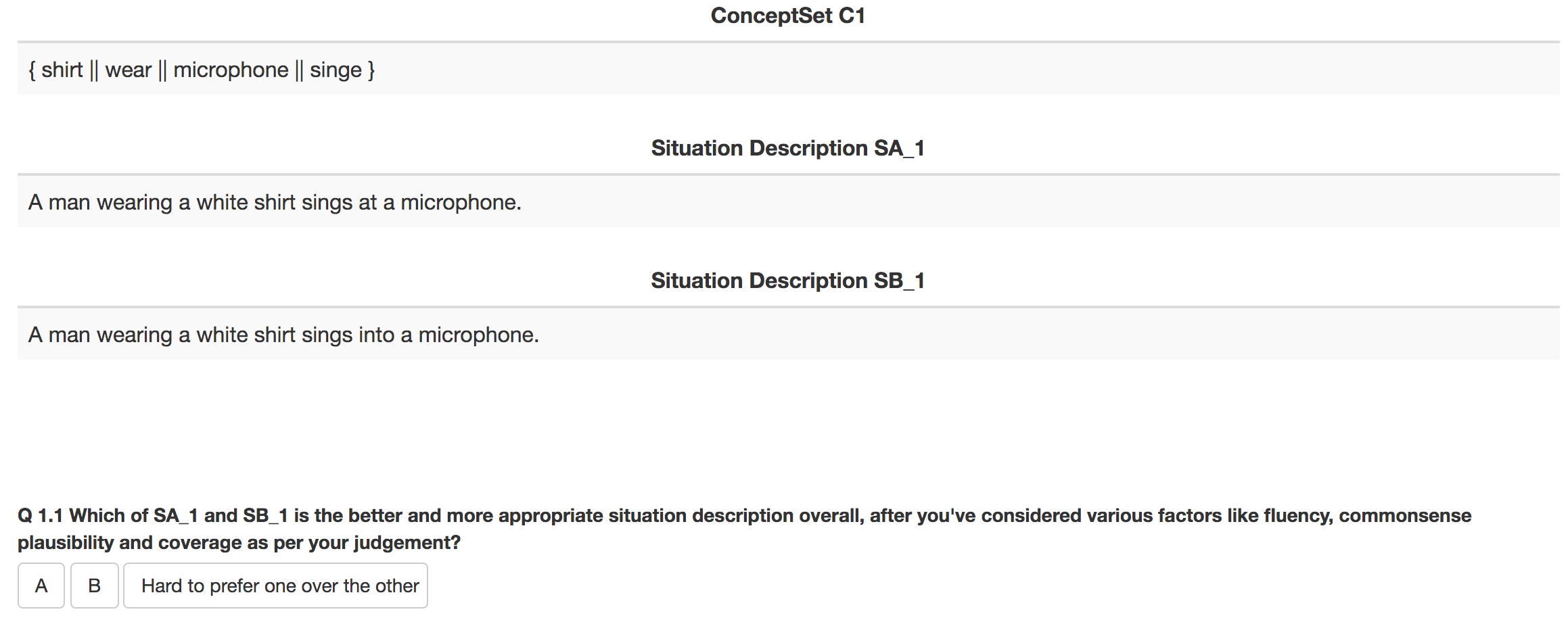}
    \caption{}
    \label{fig:amt_preserve_questions_graph}
\end{subfigure}
\caption{\small Snapshots of human evaluation: a) instructions seen by annotator and b) an example with questions.\label{fig:human_eval_template}}
\end{figure*}

\begin{table*}[ht!]
\centering
\small
\addtolength{\tabcolsep}{-4pt}
\begin{tabular}{|p{2.8cm}|p{13cm}|}
\hline 
\textbf{Method} & \textbf{Text} \\ \hline
Concept set & \{sunglass, wear, lady, sit\}\\ \hline
Captions & {a woman sitting on a bench with a cell phone} $<$s$>$ {a woman sitting on a bench with a book}\\ \hline
T5-base-BL & {A lady sits in a sunglass.} \\ \hline
T5-base-VisCTG & {A lady wearing sunglasses sits on a bench.} \\\hline 
Human reference & {The lady wants to wear sunglasses, sit, relax, and enjoy her afternoon.} \\
\Xhline{3\arrayrulewidth}
Concept set & \{music, dance, room, listen\}\\ \hline
Captions & {a person is standing in a room with a bed} $<$s$>$ {a woman is holding a laptop in a room}\\ \hline
BART-large-BL & {A listening music and dancing in a dark room} \\ \hline
BART-large-VisCTG & {A group of people are dancing and listening to music in a room.} \\ \hline 
Human reference & {A boy danced around the room while listening to music.} \\
\Xhline{3\arrayrulewidth}
Concept set & \{pool, water, slide, slide\}\\ \hline
Captions & {a boat is parked in a water with a boat}\\ \hline
T5-large-BL & {A girl slides into a pool and slides into the water.} \\ \hline
T5-large-VisCTG & {A group of people slide down a slide into a pool of water.} \\ \hline 
Human reference & {A boy slides down a bouncy slide into a pool of water.} \\
\Xhline{3\arrayrulewidth}
Concept set & \{rock, water, stand, body\}\\ \hline
Captions & {a bird sitting on a rock in a body of water}\\ \hline
T5-large-BL & {a body of water standing on rocks} \\ \hline
T5-large-VisCTG & {A man standing on a rock near a body of water.} \\ \hline 
Human reference & {A bird standing on a large rock in a body of water.} \\
\Xhline{3\arrayrulewidth}
Concept set & \{card, deck, shuffle, hand\}\\ \hline
Captions & {a person holding a cell phone in their hand} $<$s$>$ {a person holding a pair of scissors in their hand}\\ \hline
BART-large-BL & {a hand shakes a deck of cards} \\ \hline
BART-large-VisCTG & {A man shuffles a deck of cards with his hand.} \\ \hline 
Human reference & {A man shuffles a deck of cards in his hands.} \\
\Xhline{3\arrayrulewidth}
Concept set & \{chase, ball, owner, dog, throw\}\\ \hline
Captions & {a dog is standing in the grass with a frisbee} $<$s$>$ {a dog is playing with a frisbee in the grass}\\ \hline
T5-base-BL & {owner throws a ball to his dog during a chase.} \\ \hline
T5-base-VisCTG & {A dog is throwing a ball at its owner.} \\ \hline 
Human reference & {The owner threw the ball for the dog to chase after.} \\
\Xhline{3\arrayrulewidth}
Concept set & \{body, water, bench, sit\}\\ \hline
Captions & {a bench sitting on a beach next to a body of water} $<$s$>$ {a man is sitting on a bench with a cell phone} $<$s$>$ {a bench sitting on a of a beach} $<$s$>$ {a bench sitting in the middle of a lake} $<$s$>$ {woman sitting on a bench with a bird in the background}\\ \hline
BART-base-BL & {A woman sitting on a bench with water in her body.} \\ \hline
BART-base-VisCTG & {A man sits on a bench near a body of water.} \\ \hline
Human reference & {The woman sat on the bench as she stared at the body of water.} \\
\Xhline{3\arrayrulewidth}
Concept set & \{bench, sit, talk, phone\}\\ \hline
Captions & {a man sitting on a bench with a cell phone} $<$s$>$ {a woman sitting on a bench with a cell phone} $<$s$>$ {a man sitting on a bench with a cell phone} $<$s$>$ {a person sitting on a bench with a skateboard} $<$s$>$ {a man sitting on a bench with a laptop}\\ \hline
BART-base-BL & {A man sitting on a bench talking to his phone.} \\ \hline
BART-base-VisCTG & {A man sitting on a bench talking on his cell phone.} \\ \hline 
Human reference & {The woman sits on the bench to talk on her daughter on the phone.} \\
\hline \end{tabular}
\caption{\small Further qualitative examples for test$_{CG}$. \textit{BL} stands for baseline. \textit{Concept set} refers to the input keywords and \textit{Captions} refers to the {captions} (separated by $<$s$>$) used by the VisCTG model for that particular example to produce its final generation.}
\label{tab:qualitative_appendix}
\end{table*}

\section{Full Re-implementation versus Reported Model Numbers}\label{appendix:reimplementation_numbers}
See Table \ref{tab:full_reimplementation_stats} for a full comparison (across all metrics) of our re-implemented CommonGen models compared to the original reported baseline numbers in \citet{lin-etal-2020-commongen}.

\section{Pretrained FC Image Captioning Model Details}\label{appendix:image_captioning_model}
The image encoder is a pretrained Resnet-101 \cite{NIPS2015_14bfa6bb}, where the global avg. pooling of the final convolutional layer output, a vector of dim. 2048, is taken per image. Spatial features are extracted from the output of a Faster R-CNN \cite{NIPS2015_14bfa6bb,anderson2018bottomup} with ResNet-101 \cite{DBLP:conf/cvpr/HeZRS16}, trained by object and attribute annotations from Visual Genome \cite{krishna2016visual}. For captioning, the dimensions of LSTM hidden state, image feature embedding, and word embedding are all set to 512. Please see \citet{luo2018discriminability}, particularly Sections 3.3 and 5.1, and \citet{Rennie_2017_CVPR}, particularly Sections 2 and 5, for more.

\section{BART and T5 Model Training and Generation Details}\label{appendix:model_training_finetuning_details}

T5-large has 770M params, T5-base 220M params, BART-large 406M params, and BART-base 139M params. Two seeded versions of each baseline and VisCTG model are trained. For decoding, we use beam search with a beam size of 5, decoder early stopping, a decoder length penalty of 0.6, a decoder maximum length of 32, and a decoder minimum length of 1 for all models. We use a maximum encoder length of 32 for the baselines and for the VisCTG models: up to 160 for BART and 256 for T5. A batch size of 64 for T5-base and BART-base, 32 for BART-large, and 8 for T5-large is used for training. We 500 warmup steps for BART-large, and 400 for T5-base, T5-large, and BART-base. All models are trained up to a reasonable number of epochs (e.g. 10 or 20) and early stopping using our best judgment is conducted, e.g. if metrics continuously drop for several epochs. Learning rates for VisCTG models were determined by trying several values (e.g. from 1e-6 to 1e-4), and finding ones which result in decent convergence behavior, e.g. dev metrics increase steadily and reach a maximum after a reasonable number of epochs. For the final models (e.g. best NTC values for VisCTG), learning rates are (each set consists of \{BART-base,BART-large,T5-base,T5-large\}): baselines = \{3e-05,3e-05,5e-05,2e-05\}, VisCTG = \{1e-05,5e-06,2e-05,2e-05\}.

Google Colab instances were used for training, which used either a single V100 or P100 GPU. Most of the training experiments were performed using a single V100. BART-base models trained in approx. 1 hour, T5-base models in approx. 1.5 hours, BART-large models in approx. 2 hours, and T5-large models in approx. 6 hours. 

\section{Human Evaluation Details}\label{sec:appendix_human_eval} 
The Amazon Mechanical Turk (AMT) human evaluation was performed through paid annotators on AMT. Annotators were from Anglophone countries with $>97$\% approval rate. Each example was evaluated by up to three annotators. Each AMT task page or HIT contained 2 actual examples and a ``quality-check" example in random order. Specific instructions and a question snippet can be seen in Figure \ref{fig:human_eval_template}.

On every annotation page, we include one randomly chosen ``quality-check" example from a list of such hand-crafted examples, in addition to two actual examples with VisCTG and baseline outputs. The hand-crafted examples are constructed to have an obviously good and an obviously bad output pair, and are sourced from \citet{lin-etal-2020-commongen}. 
If an annotator answers the quality-check question wrong (e.g. they choose the obviously bad output), their two remaining actual example annotations are excluded while compiling results.

The time given for each AMT task instance or HIT was 8 minutes. Sufficient time to read the instructions, as calibrated by authors, was also considered in the maximum time limit for each HIT/task. Annotators were paid 98 cents per HIT. The rate of payment (\$7.35/hour) exceeds the minimum wage rate for the USA (\$7.2/hour) and hence constitutes fair pay. We neither solicit, record, request, or predict any personal information pertaining to the AMT crowdworkers.

The expert linguist evaluation included a human subject institutional board protocol and a rate of payment of \$15/hour, also exceeding the minimum wage rate for the USA.

\section{Further Qualitative Examples}\label{sec:appendix_qualitative_examples}

See Table \ref{tab:qualitative_appendix} for further qualitative examples.

\end{document}